\documentclass[12pt,letterpaper]{article}
\usepackage[a4paper, total={7in, 10in}]{geometry}

\usepackage{helvet}
\usepackage{authblk}
\usepackage{hyperref}

\makeatletter
\renewcommand{\maketitle}{\bgroup\setlength{\parindent}{0pt}
\begin{flushleft}
  \textbf{\@title}
  
  \@author
\end{flushleft}\egroup}
\makeatother

\usepackage{crop}
\usepackage{graphicx}
\usepackage{caption}
\usepackage{hyperref}
\usepackage{comment}
\usepackage[utf8]{inputenc}
\usepackage[T1]{fontenc}
\usepackage{bm}
\usepackage{hyperref}
\usepackage{color}
\usepackage{xcolor}
\usepackage{amsmath}
\usepackage{amsfonts}
\usepackage{moreverb} 
\usepackage{setspace}
\usepackage{array}
\usepackage{booktabs}
\usepackage{makecell}
\usepackage{multirow}
\usepackage[bb=dsserif]{mathalpha}
\usepackage{amssymb}

\newcommand{\supref}[1]{{\hyperref[supplementary]{Supplementary~{#1}}}}

\newcommand{\figref}[1]{{Fig.~\ref{#1}}}

\title{Metric-DST: Mitigating Selection Bias Through Diversity-Guided Semi-Supervised Metric Learning}
\date{}

\author[1,*]{Yasin I. Tepeli}
\author[1]{Mathijs de Wolf}
\author[1,*,+]{Joana P. Gon\c{c}alves}

\affil[1]{Pattern Recognition \& Bioinformatics, Department of Intelligent Systems, EEMCS Faculty, Delft University of Technology, Delft, 2628 XE, The Netherlands}
\affil[+]{The lead contact}
\affil[*]{Correspondence: \{y.i.tepeli,joana.goncalves\}@tudelft.nl}

\usepackage{cite}

\begin{document}

\maketitle

\section*{Summary}
Selection bias poses a critical challenge for fairness in machine learning, as models trained on data that is less representative of the population might exhibit undesirable behavior for underrepresented profiles. Semi-supervised learning strategies like self-training can mitigate selection bias by incorporating unlabeled data into model training to gain further insight into the distribution of the population. However, conventional self-training seeks to include high-confidence data samples, which may reinforce existing model bias and compromise effectiveness. 
We propose Metric-DST, a diversity-guided self-training strategy that leverages metric learning and its implicit embedding space to counter confidence-based bias through the inclusion of more diverse samples.
Metric-DST learned more robust models in the presence of selection bias for generated and real-world datasets with induced bias, as well as  a molecular biology prediction task with intrinsic bias. The Metric-DST learning strategy offers a flexible and widely applicable solution to mitigate selection bias and enhance fairness of machine learning models. 
\section*{Keywords}

Selection bias, Metric learning, Semi-supervised learning, Diversity, Fairness, Machine learning.
\section*{Introduction}

Machine learning (ML) algorithms enabling predictive modeling and data-driven decision-making have contributed important advances across disciplines. The increasing pervasiveness of ML in society also raises awareness about its potential impact on people's lives and the need to ensure fairness in predictions made by ML models. Selection bias is one of the most common sources of unfairness in ML, where the training data is not representative of the underlying population, with some groups or profiles appearing more prominently while others might be excluded \cite{Wu2008,Persello2014,Richards2011,Shen2022,Melucci2016}.

Mitigating selection bias is crucial to ensure fairness, accuracy, and reliability of machine learning models. Several approaches have been proposed to address this issue, including data preprocessing techniques \cite{Blitzer2006,Fernando2013,Kouw2016}, reweighting methods \cite{Zadrozny2004,Chang2009,Seah2011,Sugiyama2013,Nguyen2016,Huang2006,Du2021}, and algorithmic fairness measures \cite{Liu2014,Kouw2021tcpr}. Most of these methods are proposed under the umbrella term of domain adaptation (DA), which adjusts models to account for distribution shifts between source and target prediction domains. Available DA approaches typically focus on adapting models to specific test sets, which can limit the generalizability of the models beyond the train and test data.

Semi-supervised learning has gained traction to address bias by leveraging abundant unlabeled data that might offer further insight into the true underlying distribution of the data but cannot be directly used in supervised learning. 
A common framework for semi-supervised learning relies on self-training that iterates between (i) building a model with supervised learning and (ii) using the model both to predict pseudo-labels for unlabeled samples and to select a subset to  incorporate into the learning during the subsequent iteration. Conventional self-training selects pseudo-labeled samples based on model confidence, often focusing on the most confident predictions~\cite{vanEngelen2019,DongHyun2013}, which can reinforce the bias in the data by incorporating samples similar to others already in the biased labeled set~\cite{Radhakrishnan2023,Tepeli2024,Arazo2019}.  

To counteract this confirmation bias, the DCAST~\cite{Tepeli2024} semi-supervised strategy gradually includes diverse pseudo-labeled samples above a relaxed confidence threshold. Diversity is achieved by choosing samples from distinct clusters, identified based on sample distances or dissimilarities. The preferred DCAST approach leverages distances within a learned class-informed latent space, rather than the original feature space, to lessen the influence of uninformative features. This can be especially important for high-dimensional data, however the approach cannot be combined with classifiers lacking such latent representations. Additionally, DCAST presumes that the different clusters in the latent space capture diverse sets of samples, which can be suboptimal if the data cannot be meaningfully clustered. 

We introduce Metric-DST, a self-training framework relying on metric learning to enable more general selection bias mitigation for diversity-aware prediction models. Metric learning offers a suitable alternative to obtain a class-informed latent space~\cite{Chopra2005} by optimizing a transformation of the original feature space to a lower dimensionality in a class-contrastive manner. Metric-DST uses this mechanism to learn a bounded latent space where distances between samples reflect both dissimilarity and class membership, and then generates random locations within the space to select diverse samples that are predicted by a companion classifier above a relaxed confidence threshold. Metric-DST exploits sample diversity during model learning to improve generalizability, and can be used with virtually any type of classifier. 

\begin{figure*}[!htb]%
\centering
\includegraphics[width=\textwidth]{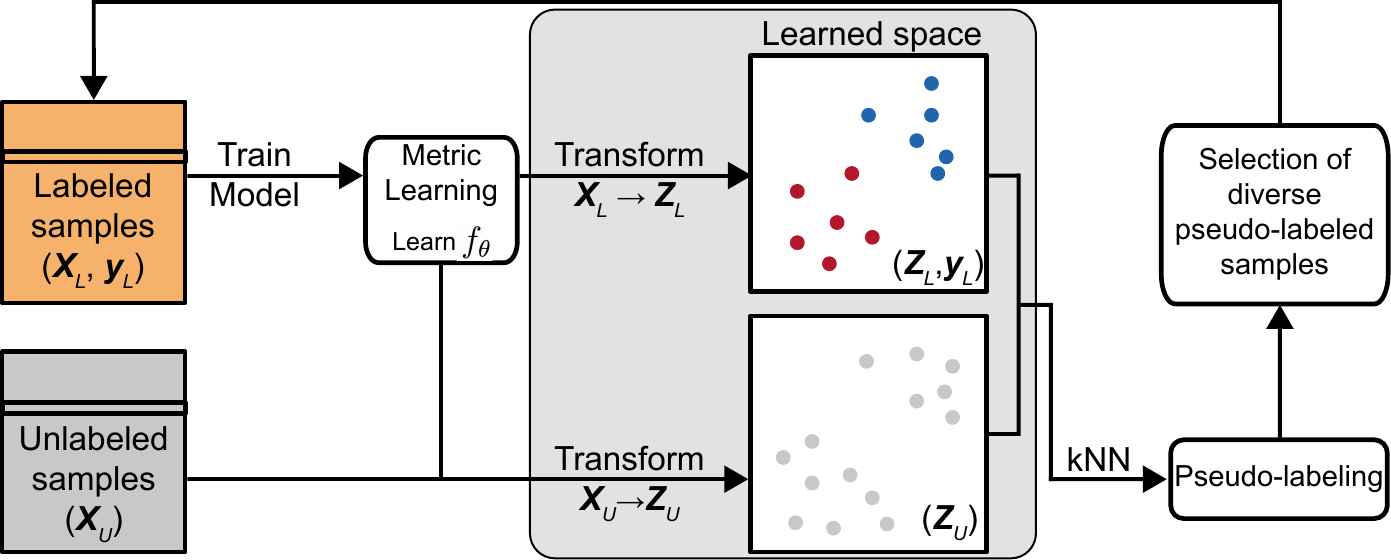}
\caption{\textbf{Overview of the Metric-DST methodology.}
A Metric-DST iteration encompasses 1) training a metric learning model on labeled data that can be used to transform both labeled and unlabeled samples into an embedding space, 2) obtaining predicted pseudo-labels and model confidence values for unlabeled samples using k-nearest neighbors (kNN) on the embedding space representations, 3) selecting diverse pseudo-labeled samples distributed across the learned embedding space 
and 
adding them to the labeled set for the subsequent iteration.}\label{fig:overview}
\vspace{-0.5cm}
\end{figure*}

\section*{Results and Discussion}
\subsection*{Model learning under selection bias with Metric-DST}
The aim of the proposed semi-supervised Metric-DST framework is to learn a prediction model with improved robustness to selection bias by leveraging available unlabeled data for additional representativeness of the underlying population distribution (\figref{fig:overview}). Using self-training, unlabeled samples are gradually pseudo-labeled and selected to be incorporated into model learning. Since conventional self-training is prone to reinforcing data bias, Metric-DST seeks to counter such behavior through the selection of diverse pseudo-labeled samples. To achieve this, Metric-DST exploits a metric learning model formulation to generate class-informative representations of samples in a bounded latent space. 
Briefly, at each self-training iteration, Metric-DST first learns a transformation function or model $f_\theta$ from the labeled samples $\boldsymbol{X}_L$ and respective labels $\boldsymbol{y}_L$ using metric learning with a contrastive loss to optimize class separation in the learned latent space (\figref{fig:overview}, \hyperref[sec:methods-metric-dst]{Methods}). The learned transformation $f_\theta$ is used to obtain embeddings or representations $\boldsymbol{Z}_U$ of the unlabeled samples $\boldsymbol{X}_U$ in the new space. 
Then, the learned representations are used by Metric-DST in two ways: (i) to make predictions and thus assign pseudo-labels to unlabeled samples, using a simple weighted $k$ nearest neighbors classifier; and (ii) to select $p/2$ diverse pseudo-labeled samples per class as randomly generated points in the latent space whose nearest pseudo-labeled sample satisfies a relaxed confidence threshold $\mu$. 

We evaluated the bias mitigation ability of the proposed diversity-guided Metric-DST method against two approaches: Metric-ST, a similarly semi-supervised variant relying on conventional self-training without diversity; and Supervised, vanilla supervised learning. Generally, our goal was to investigate if Metric-DST could build models with improved robustness to selection bias and if the diversity strategy was effective in that regard. All three strategies used metric learning with an identical neural network architecture, in combination with weighted $k$NN for prediction. Different bias scenarios were also considered across generated and real-world benchmark binary classification datasets, as well as a molecular biology challenge inherently affected by selection bias called synthetic lethality prediction. Each of the three learning methods was assessed for each bias scenario across 10 different train/test splits  (\hyperref[sec:datasets_experimentalsettings]{Methods}).

\subsection*{Metric-DST mitigates bias induced to generated and real-world datasets}
We first evaluated Metric-DST, and the Metric-ST and Supervised baselines, on binary classification tasks using artificially generated and real-world benchmark datasets with induced selection bias. Briefly, for each train/test split, the train set comprising 90\% of the data was further randomly split into labeled (30\%) and unlabeled (70\%) subsets. The Supervised approach trained using labeled data alone, while Metric-(D)ST trained using both labeled and unlabeled data. For experiments using bias, selection bias was induced only to the labeled subset, enabling us to assess if the trained model could generalize beyond the biased training data and also leverage the unlabeled data to do so. For comparison, we also trained separate models without bias induction and using a random selection of samples (as many as used in the biased selection).

\begin{figure*}[!t]%
    \centering
    \includegraphics[width=\textwidth]{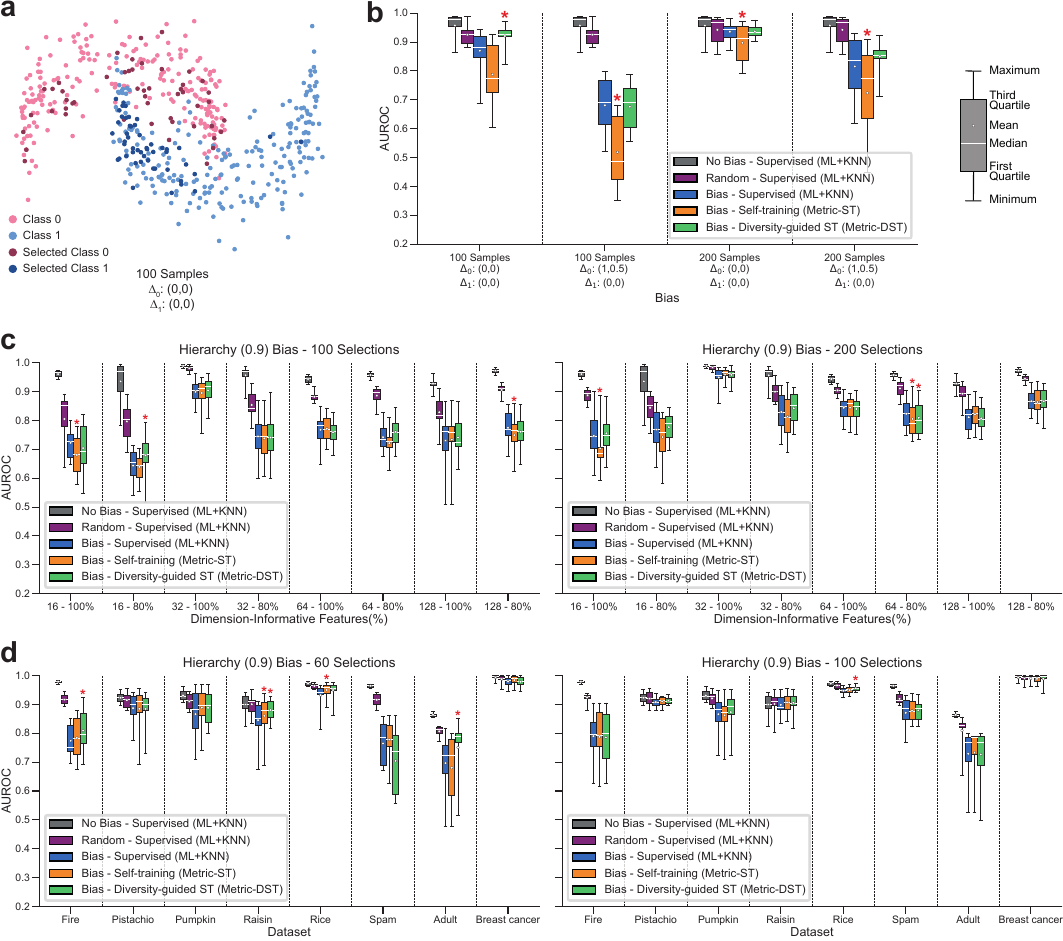}
    \caption{\textbf{Mitigation of selection bias induced to generated and real-world benchmark data.}
    \textbf{(a)} Samples selected by delta bias ($\Delta_0=\Delta_1=(0,0)$ for classes $0$ and $1$) highlighted on a scatter plot of the artificially generated 2D moons dataset. 
    Performance (AUROC) of supervised and semi-supervised Metric-(D)ST methods using metric learning and kNN on: \textbf{(b)} generated 2D moons dataset of 2000 samples with four delta bias induction settings, selecting 100 or 200 samples with $\Delta_0=\Delta_1=(0,0)$ and $\{\Delta_0=(1,0.5), \Delta_1=(0,0)\}$, \textbf{(c)} generated higher-dimensional datasets of 2000 samples and 16, 32, 64, and 128 features with hierarchy bias induction (ratio $b=0.9$) selecting 100 or 200 samples, and \textbf{(d)} eight real-worlddatasets with hierarchy bias induction ($b=0.9$) targeting the selection of 60 and 100 samples. Results of 10-fold cross-validation, with all methods evaluated using the same folds (train/test splits) and the same divisions of the train sets into labeled and unlabeled subsets. Methods included: supervised model trained on the complete labeled set (No Bias), on a biased selection (Bias), or on randomly selected samples (Random, same number as the biased selection); and semi-supervised models, using conventional self-training (Metric-ST) or diversity-guided self-training (Metric-DST) on the biased labeled train set plus the unlabeled train set. The red asterisks stand for significant difference (p-value<0.05) between the performances of the method with asterisk and the biased supervised method based on a two-sided Wilcoxon signed-rank test.}\label{fig:toy_real_datasets}
\vspace{-0.5cm}
\end{figure*}

\paragraph{Moons dataset and delta bias.} 
The generated moons dataset contained 2000 data points in 2 dimensions, distributed over two classes, with the class-specific point clouds forming interleaving moon shapes (\figref{fig:toy_real_datasets}a). We induced selection bias using a technique termed delta bias to obtain a set of either 100 or 200 class-balanced samples in the vicinity of user-defined points $\Delta_0$ and $\Delta_1$ for classes $0$ and $1$, respectively. We also used two combinations of $\Delta$ points: identical for both classes, $\Delta_0=\Delta_1=(0,0)$; and different per class, with $\{\Delta_0=(1,0.5), \Delta_1=(0,0)\}$. The effect of delta bias was confirmed by visualizing the samples in the 2D space. We observed that the biased selection excluded relevant regions of the point clouds, which could shift the decision boundary of a classifier (\figref{fig:toy_real_datasets}a). 

Selection bias had a noticeable impact on models built using supervised learning, where training on a biased selection generally resulted in lower performances compared to training on the original data (\figref{fig:toy_real_datasets}b, blue vs. grey), with differences in median AUROC between 0.02 and 0.28. 
The effect of supervised training on a biased selection was also larger than that of training on a random selection with the same number of samples (\figref{fig:toy_real_datasets}b, blue vs. purple), enabling us to disentangle the influence of bias and sample size. 
We further noticed that the decrease in supervised learning performance was stronger using selection bias with distinct $\Delta$ points per class, leading to median AUROC values of 0.69 and 0.84 for 100 and 200 samples, compared to 0.88 and 0.95 using identical $\Delta$ points (\figref{fig:toy_real_datasets}b, blue). 
The Metric-ST variant without diversity was unable to overcome the induced selection bias, leading to large variances accompanied by decreases in performance compared to supervised learning across all four bias settings (\figref{fig:toy_real_datasets}b, yellow vs. blue), three of which were statistically significant (p-values < 0.03). 
In contrast, the diversity-guided Metric-DST method showed a significant improvement in performance with 100 samples and identical $\Delta$ points (median AUROC: supervised 0.88, Metric-DST 0.93, p-value: 0.037) and no significant performance differences but smaller variances in performance for the three remaining bias settings compared to supervised learning (\figref{fig:toy_real_datasets}b, green vs. blue). 

Overall, on the moons dataset, Metric-DST delivered models with increased robustness to induced delta bias compared to conventional self-training (Metric-ST). The proposed diversity-guided approach also performed comparably or better than supervised learning. 

\paragraph{Higher-dimensional two-cluster datasets and hierarchy bias.} 
We complemented the generated data using 8 balanced binary classification datasets of 2000 samples spread over two clusters per class. The datasets spanned four dimensionalities or numbers of features (16, 32, 64, and 128), paired with an additional setting determining whether 100\% or 80\% of those features were informative for the classification task. We selected a biased subset of 100 or 200 samples from each dataset using hierarchy bias with bias ratio $b = 0.9$~\cite{Tepeli2024}, which favored samples from one specific cluster identified de novo per class (\hyperref[sec:datasets_experimentalsettings]{Methods}, Supplementary Fig. S1). 

Training on a random selection of 100 or 200 samples caused a decrease in the performance of supervised learning across 7 of the 8 datasets compared to training without bias (\figref{fig:toy_real_datasets}c, purple vs. grey). The biased selection using hierarchy bias led to a further decrease in supervised model performance beyond the impact of random selection and respective reduction in sample size, with a change in median AUROC between 0.06 and 0.17 for 100 samples and between 0.03 and 0.14 for 200 samples (\figref{fig:toy_real_datasets}c, blue vs. purple). 
The Metric-ST method relying on conventional self-training was comparable or worse than supervised learning concerning robustness to induced hierarchy bias, and led to significant decreases in performance for 2 out of the 8 datasets for both 100 and 200 selected samples (\figref{fig:toy_real_datasets}c, yellow vs. blue, p-values < 0.03). 
Metric-DST was mostly comparable to supervised learning, with only two significant differences: a performance increase for 100 samples with 16 dimensions of which 80\% informative (\figref{fig:toy_real_datasets}c, green vs. blue, p-value < 0.05), and a performance drop for the 200 sample selection of the 64-dimensional dataset with 80\% informative features (p-value 0.004). We also observed non-significant increases in median AUROC for 100 samples with 64 dimensions of which 80\% informative (medians 0.73 vs 0.76) and for 200 samples with 16 and 32 dimensions of which 80\% informative (medians 0.77 vs 0.79 for 16 dimensions and 0.83 vs 0.85 for 32 dimensions). 

On the generated higher-dimensional datasets, Metric-DST displayed superior robustness to induced hierarchy selection bias compared to the Metric-ST approach. Mostly Metric-DST was able to protect the supervised learning performance, with occasional very modest improvements. 

\paragraph{Real-world benchmark datasets with hierarchy bias.} 
Event though artificially generated data and bias induction may offer some sense of control over the conditions of the experiments, there is still a multiplicity of factors to consider, and it is unlikely that the generated datasets capture the complexity and exhibit the behavior of real-world datasets. For this reason, we also evaluated the mitigation of selection bias on 8 real-world binary classification tasks using public datasets. We induced hierarchy selection bias with ratio $b=0.9$, targeting selections of 60 and 100 samples due to the limited size of some datasets (\hyperref[sec:datasets_experimentalsettings]{Methods}, Supplementary Fig. S2). 

Training on the biased sample selection led to an overall decrease in the performance of supervised learning models compared to training on the original data or a random selection (\figref{fig:toy_real_datasets}d). The effect of the induced hierarchy bias was however less pronounced using the larger 100 sample selection, and did not significantly affect model performance for datasets like \textit{Raisin} and \textit{Breast cancer} for which the sample count corresponded to a substantial portion of the data ($\simeq$153 samples in the labeled training set before bias induction). 

Using the 60 sample selection, Metric-ST improved performance in two datasets, \textit{Raisin} and \textit{Rice} (p-values 0.006 and 0.020). Metric-DST resulted in significantly improvements for three datasets, \textit{Fire} (p-value 0.049), \textit{Raisin} (p-value 0.020), and \textit{Adult} (p-value 0.002). Additionally, Metric-DST increased performance in the \textit{Fire} dataset as well, but the change was not significant (p-value 0.064).
While Metric-ST showed potential, Metric-DST demonstrated a greater overall impact. 
Using the 100 sample selection, neither semi-supervised Metric-(D)ST approach delivered significant performance improvements consistently across datasets: only on one instance Metric-DST improved significantly over supervised learning on the biased data, on the \textit{Rice} dataset (p-value 0.020). 
It is worth noting that the larger biased selection of 100 samples did not affect the original performance as much, leaving limited room for improvement for semi-supervised learning methods. Some datasets, especially \textit{Breast cancer}, could also potentially harbor easily separable classes, a dynamic that may cause biased selections to still capture the original decision boundary, thereby rendering semi-supervised methods less effective. Overall, Metric-DST showed improved robustness to selection bias compared to Metric-ST, and the ability to preserve or improve performance compared to supervised learning across all datasets. 

\subsection*{Metric-DST mitigates selection bias for synthetic lethality prediction}
The evaluation with induced biases on generated and real-world benchmark datasets enabled us to assess the effectiveness of the learning methods in cases where the biases in the data are unknown or difficult to characterize. However, artificially induced biases also have their limitations, and the insights gained from such experiments might not translate well to real-world prediction tasks inherently affected by complex selection biases. To cover this scenario, we finally evaluated Metric-DST on a molecular biology challenge, called synthetic lethality (SL) prediction, where the set of labeled samples available for training is known to be biased. 
We performed three experiments to evaluate Metric-DST on SL prediction, which were designed to control the extent of the difference in selection bias between paired train and test sets (\hyperref[sec:datasets_experimentalsettings]{Methods}). 

\begin{figure}[!t]%
\centering
\includegraphics[width=\columnwidth]{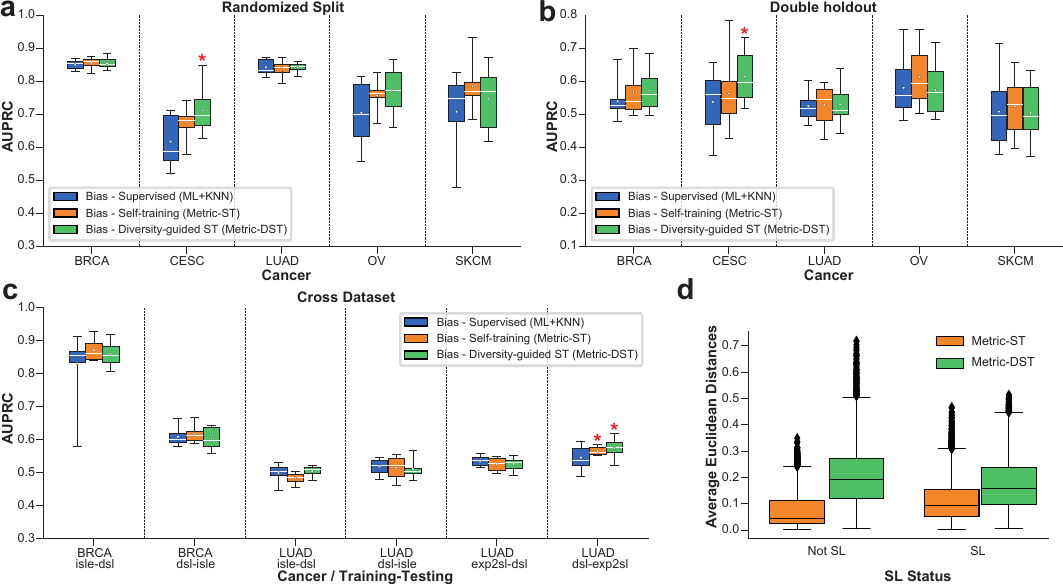}
\caption{\textbf{Mitigation of intrinsic selection bias for synthetic lethality prediction.} 
Prediction performance (AUPRC) of synthetic lethality prediction models trained and tested per cancer type using supervised learning or the semi-supervised Metric-ST and Metric-DST methods for 10 train/test splits. Three types of splits were used to control the degree of similarity in selection bias between the train and test sets: \textbf{(a)} \textit{Randomized split}, \textbf{(b)} \textit{Double holdout}, \textbf{(c)} \textit{Cross dataset}. For \textbf{(a)}, \textbf{(b)}, and \textbf{(c)}, boxplots include all points (no outlier detection), and the white circles denote the mean values. \textbf{(d)} Average Euclidean distances between pseudo-labeled samples selected by Metric-ST and Metric-DST per class, with diamonds denoting outliers. The red asterisks denote significant differences in performance (p-value <  0.05) between the method with an asterisk and the biased supervised method based on a two-sided Wilcoxon signed-rank test.} \label{fig:sl_experiments}
\vspace{-0.5cm}
\end{figure}

\paragraph{Randomized split for similar train/test selection bias.}
We assessed the supervised and semi-supervised learning methods on SL prediction for each of five distinct cancer types under similar selection bias between train and test sets. The supervised model showed noteworthy median AUPRC performances for the BRCA and LUAD cancer types (0.854 and 0.837, respectively). Metric-ST and Metric-DST both led to marginal, non-significant improvements in median AUPRC performance compared to supervised learning for LUAD (0.843 and 0.851), and Metric-DST also for BRCA (0.859) (\figref{fig:sl_experiments}a, green vs. blue). Possibly due to the ample sample sizes (Supplementary Tables S1-S3) and high starting performances of BRCA (1443 SL, 1010 non-SL pairs) and LUAD (594 SL, 5509 non-SL pairs), the use of additional pseudo-labeled data yielded inconsequential performance gains.

We noticed improvements of Metric-ST and Metric-DST over supervised learning in median AUPRC for the cancer types with more limited numbers of labeled samples, including CESC, OV, and SKCM. 
However, owing to relatively large variances, the only significant improvement was seen with Metric-DST for CESC (\figref{fig:sl_experiments}a, green vs. blue, p-value 0.014). Additionally, both Metric-DST and Metric-ST seemed superior to supervised learning for CESC and OV in median AUPRCs (Metric-DST CESC: 0.696, OV: 0.772; Metric-ST CESC: 0.683, OV: 0.765; supervised CESC: 0.587, OV: 0.701). 
We also saw a moderate non-significant improvement in median AUPRC of the Metric-(D)ST methods over supervised learning for the SKCM dataset (Metric-DST 0.769, Metric-ST 0.771, supervised 0.747). 

In summary, the application of Metric-DST looked cautiously promising in the context of a randomized split, preserving similar biases between train and test sets, for cancer types with more limited sample sizes (CESC, OV, and SKCM).

\paragraph{Double holdout for distinct train/test selection bias.} 
We also assessed Metric-DST with paired train and test sets yielding different biases, adopting a double holdout technique where gene overlap between test and train sets was entirely prevented. This restrictive split resulted in a diminished train set size, reaching its lowest for the CESC dataset with only 90 samples.

Relative to the randomized split experiment, supervised learning using double holdout resulted in lower median AUPRCs (\textit{Randomized split} vs. \textit{Double holdout} in BRCA, OV, CESC, SKCM, and LUAD: 0.853 vs. 0.527, 0.701 vs. 0.558, 0.587 vs. 0.560, 0.747 vs. 0.497, and 0.837 vs. 0.517, respectively) (Fig. \ref{fig:sl_experiments}b). This was expected due to the restrictions imposed by the double holdout to ensure zero overlap in individual genes, in addition to zero overlap in gene pairs between train and test sets. Although some performance differences could be observed between the Metric-(D)ST methods and supervised learning for the BRCA, LUAD, OV, and SKCM datasets, none of them reached statistical significance. Metric-ST showed higher median AUPRC than Metric-DST and supervised learning for LUAD, OV, and SKCM, while Metric-DST did better in this regard for BRCA and CESC. 
The only significant improvement in AUPRC performance was  recorded for CESC, with Metric-DST outperforming the supervised model (median AUPRC 0.60 vs. 0.56, p-value 0.010). It is important to note that the semi-supervised methods did not cause significant decreases in performance relative to supervised learning.

Multiple factors might explain the lack of effectiveness of Metric-DST for some cancer types. For instance, the  restrictions imposed by the double holdout procedure may have caused too extreme differences in biases between the train and test sets, due to the absence of shared genes. An additional contributing factor could be the reduction in train set size, exemplified by the CESC dataset (Supplementary Table S4). The impact of these constraints also resulted in a large performance decrease for the baseline supervised model (Fig. \ref{fig:sl_experiments}a-b), making the recovery more difficult for the semi-supervised techniques which rely  heavily on an initial successful model.

\paragraph{Cross dataset split with naturally occurring selection bias.} 
To evaluate bias mitigation with naturally occurring differences in selection bias between train and test sets, we set up the data splits to train using SL labeled samples from one study and test on SL labeled samples from another study, encompassing six permutations across three studies (ISLE, dSL, and EXP2SL). 

For BRCA, when trained on ISLE and tested on dSL, both Metric-ST and Metric-DST induced an increase in the minimum AUPRC performance by over 0.2, but overall there were no significant differences in performance between the two semi-supervised methods and supervised learning (Fig. \ref{fig:sl_experiments}c).  
For LUAD, the Metric-(D)ST methods resulted in significant performance improvements only for the setting that trained on dSL and tested on EXP2SL significant differences (median AUPRC: Supervised 0.536; Metric-ST 0.561 with p-value 0.049; Metric-DST 0.576 with p-value 0.014). The remaining study combinations did not reveal significant changes either, but we observed small decreases in median AUPRC for Metric-ST trained on ISLE and tested on dSL, as well as for Metric-DST trained on dSL and tested on ISLE.

Taking all experiments on synthetic lethality prediction into account, it is important to highlight that the two semi-supervised Metric-(D)ST methods significantly outperformed supervised learning on three scenarios, while never performing significantly worse. Instances where Metric-ST and Metric-DST yielded no clear impact might be attributed to multiple factors, including the inherent complexity of the problem with baseline supervised learning performances hovering around 0.5, or extreme disparities between the train and test sets.

\begin{figure}[!t]%
\centering
\includegraphics[width=\columnwidth]{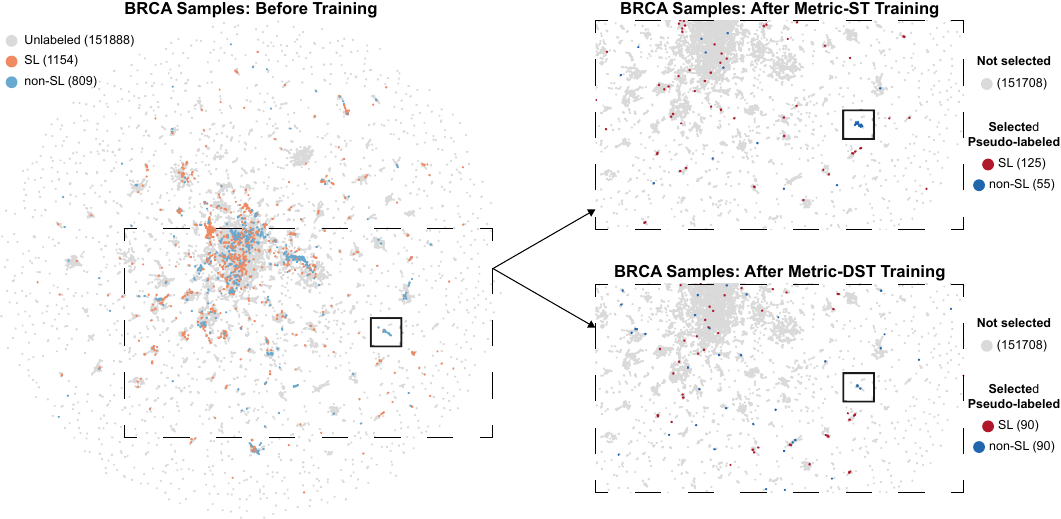}
\caption{\textbf{UMAP projections of the SL dataset for BRCA.} 
On the left, the training samples are highlighted before the training. The top right plot shows the pseudo-labeled samples selected by Metric-ST and the pseudo-labeled samples selected by Metric-ST during the training. The number of samples of each class is stated in parentheses. The highlighted box highlights a cluster dominated by gene pairs containing the gene CDH1.
} \label{fig:umap_sl}
\vspace{-0.5cm}
\end{figure}

\subsubsection*{Metric-DST promotes diversity in selected pseudo-labeled samples}
To verify if the diversity approach of Metric-DST was able to select more diverse samples, we analyzed the Euclidean pairwise distances between pseudo-labeled samples assigned to the same class label in the learned embedding space, using the BRCA \textit{Randomized split} as an example (Fig. \ref{fig:sl_experiments}d). The distances were larger on average for pseudo-samples selected by Metric-DST, confirming a more heterogeneous sample selection compared to Metric-ST. 

We also examined the distribution of selected pseudo-labeled samples in a UMAP projection of the labeled and unlabeled samples original feature space onto two dimensions (Fig. \ref{fig:umap_sl}). The projection showed no clear linear separation between the labeled samples of the two classes, SL and non-SL, reflecting the complexity of the prediction task and its underlying decision boundary. In addition, most clusters apparent in the UMAP embedding contained no labeled samples that could be used for supervised training, which further illustrates the lack of representation and the extent of the selection bias problem in synthetic lethality. 

More detailed analysis revealed that Metric-ST incorporated a total of 55 and 125 pseudo-labeled samples, respectively assigned non-SL and SL labels. Of the 55 non-SL pseudo-labeled samples, 29 were in a cluster dominated by the gene CDH1 (Fig. \ref{fig:umap_sl}). This cluster originally contained 507 labeled and unlabeled samples, of which 501 contained the gene CDH1. The fact that the method focused heavily on one cluster demonstrates the main drawback of using  conventional self-training and relying on model confidence alone to mitigate the effect of selection bias. In contrast, Metric-DST was able to select a more varied set of 90 pseudo-labeled samples, of which only 6 originated from the ``CHD1'' cluster. 

Together, these findings highlight the ability of Metric-DST to promote diversity while incorporating unlabeled samples into the learning of a prediction model. 

\section*{Conclusion}
In this work, we proposed Metric-DST, a semi-supervised  framework coupled with metric learning to build prediction models with improved robustness to sample selection bias. Metric-DST relies on self-training to incorporate unlabeled samples for additional representation and insight into the underlying distribution of the population. Crucially, Metric-DST introduces a strategy to counter confirmation bias of conventional self-training by learning from a more diverse set of samples. Diversity is introduced via metric learning of a class-contrastive representation, which facilitates the pseudo-labeling and identification of dissimilar unlabeled samples to include in the training. 

Evaluation using artificially generated and real-world datasets with induced selection bias suggested the potential of self-training to enhance model generalizability, yet also its susceptibility to exacerbate data bias. The proposed diversity-guided approach, Metric-DST, showed greater resilience than conventional self-training, albeit with modest performance improvements. 
Application to synthetic lethality prediction showed that semi-supervised metric learning could augment performance in scenarios where train and test sets yielded similar or distinct naturally occurring selection biases. It was reassuring that Metric-DST was able to preserve the performance obtained with supervised learning or deliver more robust models in all contexts, and especially under challenging conditions, such as with limited numbers of training samples or weak baseline models. 

Utimately, the effectiveness of Metric-DST is contingent upon factors such as the performance of the underlying base model, the type and extent of the data bias, and the ratio of features to samples, among others. Future work warrants a deeper exploration of the potential of the Metric-(D)ST learning framework, including refinement of neural network architectures and loss functions. Leveraging metric learning as a means of diversifying pseudo-sample selection in combination with various classifiers could further expand the scope of the model. We also envision further addressing the limitations of the existing pseudo-labeled sample selection approach, which could be extended to ensure a more comprehensive representation of the embedding space by excluding unpopulated regions.

\section*{Experimental procedures}
\label{sec:methods}
\subsection*{Metric-DST}
\label{sec:methods-metric-dst}
Metric-DST is a semi-supervised ML framework based on metric learning to obtain an embedding function or transformation that is informative for a classification task of interest with increased robustness to selection bias. Learning is accomplished via self-training, where the transformation is gradually refined by incorporating a diverse selection of newly pseudo-labeled unlabeled examples into the training process. The learned transformation serves the dual purpose of predicting pseudo-labels and assessing sample diversity to counter the data bias.

Each self-training iteration involves three steps: (1) learn a metric embedding function from the labeled data such that the latent representation of a sample also yields pertinent information about class separation, (2) pseudo-label unlabeled samples based on the learned transformation so they can be considered as candidates for selection and training, (3) select a diverse subset of pseudo-labeled samples and include them in the labeled train set for the next iteration.

\subsubsection*{Learning of a metric embedding function}

At iteration $t$, Metric-DST first learns a transformation function or model $f_{\theta}^{(t)}$ based on the labeled samples in matrix $\boldsymbol{X}_L^{(t)}$ and the corresponding binary labels $\boldsymbol{y}_L^{(t)}$ using metric learning. The general goal is to learn a transformation of an individual sample vector $\boldsymbol{x}$ to a latent embedding representation $\boldsymbol{z} = f_{\theta}^{(t)}(\boldsymbol{x})$, guided by class assignments and inter-sample distances, such that samples of the same class are closer together and samples from different classes are distanced further apart in the learned embedding space. Various model architectures could be used for the transformation, in this case we used a feed-forward neural network with a single hidden layer. The model is optimized based on the contrastive loss function designed to minimize intra-class distances and maximize inter-class distances of samples in the embedding space (Eq. \ref{eq:loss}).
\begin{equation}
    \mathcal{L}_{contrastive} = \sum_{(i, j)\in P} \mathbb{1}_{y_i=y_j}\max\{0, d_{i, j} - m_{pos}\} 
    + \mathbb{1}_{y_i\neq y_j}\max\{0,m_{neg} - d_{i, j}\}
\label{eq:loss}
\end{equation}
Here, $d_{i, j}$ denotes the Euclidean distance between samples $\boldsymbol{x}_i$ and $\boldsymbol{x}_j$ in the embedding space, thus $d_{i, j} = d(f_\theta^{(t)}(\boldsymbol{x}_i), f_\theta^{(t)}(\boldsymbol{x}_j))$. Symbol $P$ represents the set of all sample pairs within a training batch, and the indicator function $\mathbb{1}_{condition}$ takes value 1 if the condition holds or 0 otherwise. The positive and negative margins, $m_{pos}$ and $m_{neg}$, are used to prevent the algorithm from forcing samples with the same labels to overlap completely or samples with different labels to be separated infinitely. Specifically, the distance between samples with the same labels only increases the loss when it exceeds the positive margin, and the distance between samples with different labels stops contributing to the loss once the distance exceeds the negative margin.

Once the transformation has been learned from the labeled samples $\boldsymbol{X}_L^{(t)}$, it can be applied to obtain embedding representations for unlabeled samples in $\boldsymbol{X}_U^{(t)}$ as well. We denote the embedding matrix containing the representations of all samples, labeled and unlabeled, by $\boldsymbol{Z}^{(t)}$. 

\subsubsection*{Pseudo-labeling of unlabeled samples through metric embedding}
\label{sec:predicting_label}
The transformation model $f_\theta^{(t)}$ learned from the labeled data cannot be directly used to make predictions and thus assign pseudo-labels to unlabeled samples. To classify the unlabeled samples, Metric-DST applies a weighted version of $k$ nearest neighbors (kNN) to the embedding matrix $\boldsymbol{Z}^{(t)}$ with the learned representations $\boldsymbol{z}^{(t)} = f_{\theta}^{(t)}(\boldsymbol{x})$ of all samples. For a given unlabeled sample $i$ with representation $\boldsymbol{z}_i^{(t)} \in \boldsymbol{Z}^{(t)}$, Metric-DST identifies the set $N_i^{(t)}$ of its $k$ closest labeled samples in $\boldsymbol{Z}^{(t)}$. The prediction class probability $\bar{y}_i \in [0,1]$ for sample $\boldsymbol{x}_i$ is then calculated as a weighted average of the probabilities of the $k$ neighbors, as given by Eq.~\ref{eq:confidence}. The calculation factors in the distance of each neighbor representation to $\boldsymbol{z}_i$, so that closer neighbors contribute more than farther ones.

\begin{equation}
\bar{y}_i = \frac{\sum_{n \in N_i^{(t)}} y_n \times (1-d_{i, n}) + (1 - y_n) \times d_{i, n}}{|k|}
\label{eq:confidence}
\end{equation}
The probability $\bar{y}_i$ represents the confidence of the model, where values close to $1$ and $0$ indicate high confidence in predicting class 1 and class 0, respectively. The final class label $\hat{y}_i$ is obtained by thresholding the probability value $\bar{y}_i$ as per Eq.~\ref{eq:classification}.

\begin{equation}
\hat{y}_i =
\begin{cases}
1, & \text{if } \bar{y}_i > 0.5\\
0, & \text{otherwise}
\end{cases}\label{eq:classification}
\end{equation}

\subsubsection*{Selection of diverse pseudo-labeled samples}
After assigning pseudo-labels, Metric-DST selects which newly pseudo-labeled samples to include in the labeled set for the subsequent training iteration. 

Conventional self-training (ST) typically chooses the $p$ newly pseudo-labeled samples with the highest confidence~\cite{DongHyun2013}, where $p$ is a user-defined parameter. The reliance on confidence alone promotes confirmation bias, where the model is likely to follow and strengthen the selection bias present in the labeled data. Additionally, ST is not class-aware in that it does not consider that the model may not be similarly confident about prediction of different classes, which could further lead to unwanted biases such as class imbalance. 
To address both issues, Metric-DST performs diversity-guided self-training (DST), which introduces sample diversity and class balancing into the selection of pseudo-labeled samples using the learned metric embedding. Diversity is achieved through randomness in the choice of each pseudo-labeled sample as follows. First, Metric-DST creates a candidate point in learned embedding space $\boldsymbol{Z}^{(t)}$ as a tuple of randomly generated coordinates in the range $[0, 1]$. Then, the pseudo-labeled sample closest to the candidate point is identified based on the Euclidean distance (Eq.~\ref{eq:confidence}). The selected pseudo-labeled sample is designated for inclusion in the labeled train set for the subsequent iteration if the confidence on its prediction surpasses a predefined relaxed threshold $\mu$. Class balance is achieved by selecting $p/2$ positive and $p/2$ negative pseudo-labeled samples sequentially using the aforementioned procedure for each self-training iteration. If Metric-DST fails to secure a sufficient number of pseudo-labeled samples within $50 \times p$ attempts for any one self-training iteration, undersampling of the majority class is employed to obtain a class balanced set of pseudo-labeled samples.

\subsection*{Evaluation of Metric-DST}
\label{sec:datasets_experimentalsettings}
We evaluated the Metric-DST semi-supervised model learning strategy proposed to mitigate sample selection bias against two baselines: Metric-ST, also a semi-supervised approach based on metric learning to train models using both labeled and unlabeled data, but paired with conventional self-training and thus missing the class-awareness and diversity elements of Metric-DST; and Supervised, referring to the traditional supervised metric learning technique to train models from labeled data alone. 
We used the same neural network model architecture as a basis with all learning strategies, consisting of an 8-dimensional hidden layer and a 2-dimensional output layer. Unless otherwise specified, the batch training size was set to 64, and the confidence threshold $\mu$ was set to 0.9. 
We further relied on weighted kNN with $k=5$ to make predictions based on the metric embedding of a sample. 
Finally, we assessed the bias mitigation ability of Metric-DST across a range of binary classification tasks and selection bias scenarios, ranging from artificially generated and real-world benchmark data with induced selection bias to an important prediction task in molecular biology intrinsically affected by selection bias. 

\subsubsection*{Datasets and selection bias}

\paragraph{Generated 2-dimensional moons dataset and induced delta bias.} 
We generated the simplest ``moons'' dataset as a binary class-balanced set of 2000 samples or points in a 2-dimensional space, such that the samples of the two classes formed interleaving half circles (or moons), using the \textit{make\_moons} function from scikit-learn~\cite{scikit-learn}. 
Selection bias was induced by choosing an equal number of samples from each class, while favoring samples closer to a point in space with user-defined coordinates $\Delta_i$ for each class $i$. We refer to this type of bias as delta bias, where we set the selection probability of each sample $\boldsymbol{x}$ according to its distance to the point $\Delta_{class(\boldsymbol{x})}$ associated with the corresponding class label $class(\boldsymbol{x})$, and then selected samples without replacement based on their normalized selection probabilities. The selection probability of a sample $\boldsymbol{x}$ was defined to decrease exponentially with the Manhattan distance to $\Delta_{class(\boldsymbol{x})}$, multiplied by a factor of 2 denoting bias strength: 
$P_{\boldsymbol{x}} = e^{-2 \times \left(|{x_1}-\Delta_{class(\boldsymbol{x}),1}| + |{x_2}-\Delta_{class(\boldsymbol{x}),2}|\right)}$, 
where $x_1$ and $x_2$ are the coordinates of $\boldsymbol{x}$ and $\Delta_{class(\boldsymbol{x}),1}$ and $\Delta_{class(\boldsymbol{x}),2}$ are the coordinates of $\Delta_{class(\boldsymbol{x})}$ in the 2D space, respectively. Four different biased selections of the moon dataset were generated, two of 100 samples and two of 200 samples, combined with ${\Delta_0 = (0, 0)}$ and ${\Delta_1 = (0,0)}$ or ${\Delta_0 = (1,0.5)}$ and ${\Delta_1 = (0,0)}$.

\paragraph{Generated higher-dimensional datasets and induced hierarchy bias.} 
\label{sec:methods-datasets-bias}
We created 8 $n$-dimensional datasets, each containing 2000 samples with binary class-balanced labels and forming two sample clusters per class, using the \textit{make\_classification} function from scikit-learn~\cite{scikit-learn}. 
Each $n$-dimensional dataset was generated with $f$ of the $n$ dimensions independent and informative for the prediction task, and the remaining $n-f$ dimensions as linear combinations of the $f$ informative features. Briefly, the procedure for the informative dimensions creates a $f$-dimensional hypercube with sides measuring 3 units, then generates clusters of samples distributed around the vertices of the hypercube (within 1 standard deviation), and finally assigns two randomly chosen clusters to each class. The $n-f$ additional dimensions are generated by linearly combining randomly selected informative features. We generated 8 datasets spanning four dimensionality values (16, 32, 64, and 128), combined with 80\% or 100\% of informative features. 
We induced selection bias using hierarchy bias \cite{Tepeli2024}, a multivariate technique which identifies clusters of samples and makes a biased selection of $k$ samples per class, where a bias ratio parameter $b$ is used to skew the representation of samples selected from one specific cluster relative to the others. To achieve this, hierarchy bias performs agglomerative hierarchical clustering until it obtains one cluster with at least $k$ samples, and then selects $k\times b$ samples uniformly at random from such cluster plus $k \times (1-b)$ samples uniformly at random from the remaining data. For the experiments with generated high-dimensional datasets, we used a challenging hierarchy bias with ratio $b = 0.9$ to create two biased selections of 100 and 200 class-balanced samples. 

\paragraph{Real-world binary classification benchmark datasets and induced hierarchy bias.} 
We used 8 publicly available binary classification benchmark datasets of varying dimensions, feature types, and complexity: 5 from the UCI Data Repository~\cite{Dua2017} (breast cancer, adult, spam, raisin, rice) 
and 3 from other sources including pistachio~\cite{Ozkan2021_pistachio}, fire~\cite{Koklu2021_fire}, and pumpkin~\cite{Koklu2021_pumpkin}. 
To induce selection bias, we again used hierarchy bias~\cite{Tepeli2024} with bias ratio $b = 0.9$ to create two biased selections of 60 and 100 class-balanced samples per dataset. The numbers of selected samples were chosen to be feasible and consistently applied across all real-world benchmark datasets. 

\paragraph{Synthetic lethality dataset and inherent selection bias.} 
\label{sec:sl_use_cases}
To assess the bias mitigation ability of Metric-DST on a real-world prediction task inherently affected by selection bias, we focused on the molecular biology challenge of synthetic lethality prediction. Synthetic lethality refers to a relationship between two genes, relevant for cancer therapy~\cite{Fong2009,Hutchinson2010}, whereby the loss-of-function of both genes leads to cell death but loss-of-function of either gene independently is not lethal\cite{Chan2011}. Computational prediction of synthetic lethality (SL) gene pairs is key to generate promising candidates for the discovery of new SL relationships. However, the existing labeled gene pairs used for training SL prediction models suffer from extensive selection bias~\cite{Seale2022}, as they are often limited to specific disease-related genes, gene families, or pathways~\cite{Jacquemont2012,Toledo2015,Etemadmoghadam2013,Hubert2013,Kranz2014}. 

Following recent work on supervised SL prediction models ELISL~\cite{Tepeli2023}, we represented each sample or gene pair by a 128-dimensional vector expressing a relationship between the embedding representation vectors of the two genes, based on amino acid sequence. This formulation was introduced to reflect the functional similarity of a pair of genes, and emerged as the most successful predictor of SL in ELISL models. We used SL labeled samples from 5 different cancers~\cite{Tepeli2023}: breast (BRCA), lung (LUAD), ovarian (OV), skin (SKCM), and cervix (CESC) (Supplementary Table S1). 
In addition to the labeled SL gene pairs, we used a set of unlabeled samples comprising pairwise combinations of 572 genes involved in cancer and DNA repair pathways, excluding any samples already present in the labeled set~\cite{Tepeli2023} (Supplementary Table S1). We did not use bias induction techniques with SL data, since the goal of this particular use case was to assess the behavior of the different model learning strategies in the presence of naturally occurring selection bias. We leveraged such bias for evaluation as described below.

\subsubsection*{Training and evaluation of prediction models}

\paragraph{Generated and real-world binary classification tasks.} 
We trained and evaluated all models using 10-fold cross-validation (CV), stratified by class. The CV procedure generated a split into train set (90\%) and test set (10\%) for each fold, with the train set further split randomly into labeled (30\%) and unlabeled (70\%) subsets. 
Supervised metric learning models were trained per fold on the corresponding labeled train subset, as well as biased and random selections of it. Metric-DST and Metric-ST were used to learn models per fold from the corresponding labeled train subset, as well as its biased and random selections, together with the unlabeled train subset. For the Metric-(D)ST methods, the number $p$ of selected pseudo-labeled samples was set as the greatest even integer smaller than or equal to $\sqrt n$, with $n$ referring to the number of labeled samples available for training. 
We induced selection bias to the labeled train subset using either delta or hierarchy bias, depending on the dataset, as previously described.
Each trained model was evaluated on the unbiased test set of the corresponding fold for which it was learned using the area under the receiver-operating characteristic curve (AUROC) as performance metric. The same folds and train set splits were used across all experiments.
We tested the significance of performance differences between the supervised model learned from biased data and Metric-(D)ST using two-sided Wilcoxon signed rank tests and a p-value threshold of 0.05.

\paragraph{Synthetic lethality prediction.} 
We evaluated Metric-DST, Metric-ST, and supervised metric learning for SL prediction with three experiments, each involving 10 runs of model training and evaluation based on different train/test splits. We largely followed an experimental setup previously proposed and refined to assess robustness to selection bias in SL prediction~\cite{Seale2022, Tepeli2023}.

The \textit{Randomized split} experiment assessed SL prediction performance without explicitly evaluating bias effects: the labeled gene pairs were randomly split into 20\% train and 80\% test data per run, with both subsets then expected to exhibit similar biases (Supplementary Table S2 for the distribution of classes). 
The two other experiments evaluated the ability of the model learning strategies to mitigate selection bias in training data. The \textit{Double holdout} split was set up to promote distinct biases between train and test data by distributing the labeled gene pairs into disjoint train/test sets per run, but this time also enforcing zero overlap of individual genes in addition to no overlap in gene pairs (More details in Supplementary Methods). 
For the \textit{Cross dataset} experiment, we took advantage of  the fact that different SL studies focus on distinct sets of genes and thus naturally yield varying selection bias. We therefore split the labeled gene pairs based on the three SL studies from which they were obtained: ISLE~\cite{Lee2018}, dSL~\cite{DiscoverSL}, EXP2SL\cite{Exp2sl}. 
Considering only cancer types and studies with a sufficient number of samples, models were trained using labeled pairs from one study and tested on labeled pairs from another study.
Any gene pairs overlapping between the train and test sets, due to their inclusion in multiple studies, were removed from the train set. 

For all three experiments, train and test sets were class-balanced at the start of each run by randomly undersampling the majority class, and 20\% of the train set was used as a validation set for early stopping (Supplementary Tables S3-S5 for the number of samples in each experiment). Each model was trained until the validation loss did not decrease for five consecutive rounds of self-training, with the final performance evaluated on the test set.
We measured performance using the area under the precision-recall curve (AUPRC) score, given that SL prediction places a greater emphasis on detecting positive SL pairs and negative pairs (non-SL) cannot be confidently identified or validated. The AUPRC score is suitable for measuring performance in this scenario, as it does not take correctly predicted negatives into account. We assessed the significance of performance differences in SL experiments using two-sided Wilcoxon signed ranked tests and a p-value significance threshold of 0.05.

The hyperparameters of Metric-(D)ST, namely the confidence threshold $\mu$ and number of pseudo-labeled samples $p$ to select per iteration, could be set judiciously for the application to other datasets using controlled bias induction. Since the effect of these hyperparameters could be more challenging to predict for the synthetic lethality dataset with inherent selection bias, we performed grid search to identify the hyperparameter values leading to the lowest validation loss per run for each experiment (Supplementary Tables S6-S8). The final performance was obtained on the test set using the model with the selected hyperparameter values.

\subsection*{Resource availability}
    \subsubsection*{Lead contact}
        Corresponding author: \href{mailto:joana.goncalves@tudelft.nl}{joana.goncalves@tudelft.nl}
    \subsubsection*{Data and code availability}
        The data used in this article were obtained from publicly available sources, detailed in the Experimental procedures section. The raw data necessary to reproduce the experiments are accessible via Figshare at \href{https://doi.org/10.6084/m9.figshare.27720726.v2}{10.6084/m9.figshare.27720726.v2}. 
        An implementation of the dataset generation, bias induction, and Metric-DST method in Python has been made available under an open source license at \href{https://github.com/joanagoncalveslab/Metric-DST}{github.com/joanagoncalveslab/Metric-DST}.

\section*{Supplemental information}
    \begin{description}
      \item Figures S1-S2, Tables S1-S8 along with their their captions and supplementary methods in a PDF
    \end{description}

\section*{Acknowledgements}
    The authors received funding from the US National Institutes of Health [U54EY032442, U54DK134302, U01DK133766, R01AG078803 to J.P.G.]. Authors are solely responsible for the research, the funders were not involved in the work. 
    The authors further acknowledge the High-Performance Compute (HPC) cluster of the Department of Intelligent Systems at the Delft University of Technology.\\\\

\section*{Author contributions}
    Conceptualization, Y.I.T., and J.P.G.; Methodology, Y.I.T., M.d.W, and J.P.G.; Validation and Formal Analysis, Y.I.T. and M.d.W; Software, Y.I.T. and M.d.W; Investigation, Y.I.T., M.d.W, and J.P.G.; Writing – Original Draft, M.d.W and Y.I.T.; Writing – Review \& Editing, J.P.G.; Funding Acquisition and Supervision, Y.I.T. and J.P.G.  \\\\

\section*{Declaration of interests}
    The authors declare no competing interests.


\end{document}


\flushbottom

\thispagestyle{empty}
\tableofcontents

\newpage
\section{Supplementary Figures}

\begin{figure*}[!h]%
    \centering
    \includegraphics[width=\textwidth]{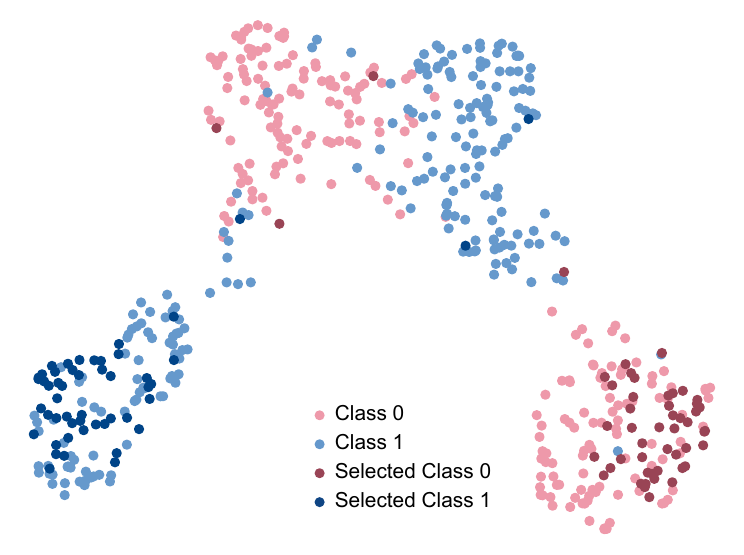}
    \caption{\textbf{Impact of hierarchy bias on the UMAP latent space for a generated higher dimensional dataset.} 100 Samples selected by hierarchy (0.9) bias highlighted on the latent UMAP space of the labeled train set for artificially generated higher dimensional dataset with 16 dimensions and 80\% informative features. Results are shown for run 1 (arbitrarily chosen).} \label{fig:umap_highdimension_run0}
\vspace{-0.5cm}
\end{figure*}

\begin{figure*}[!h]%
    \centering
    \includegraphics[width=\textwidth]{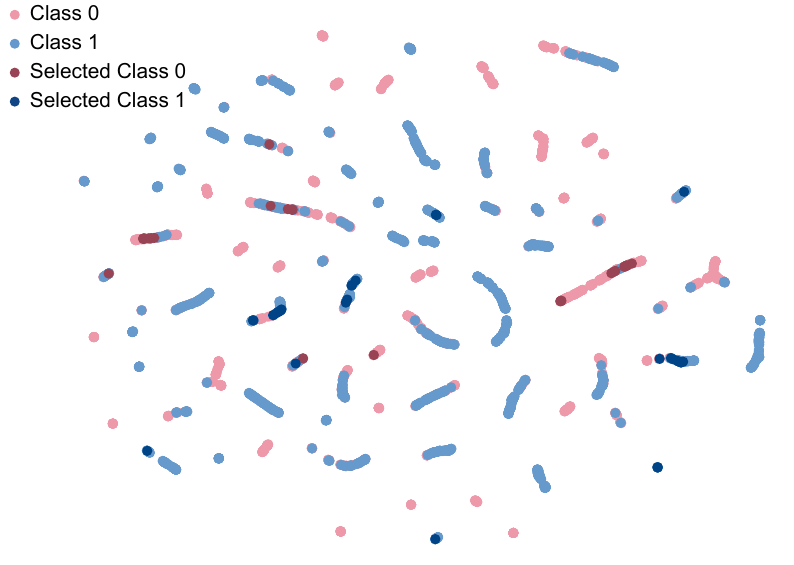}
    \caption{\textbf{Impact of hierarchy bias on the UMAP latent space for a real world fire dataset.} 60 Samples selected by hierarchy (0.9) bias highlighted on the latent UMAP space of the labeled train set for fire. Results are shown for run 7 (arbitrarily chosen).} \label{fig:umap_fire_run6}
\vspace{-0.5cm}
\end{figure*}

\newpage
\section{Supplementary Tables}
\begin{table}[!htbp]
    \centering
    \caption{Numbers of synthetic lethality labeled and unlabeled samples or gene pairs per cancer type.\label{tab:dataset_description}}
    \begin{tabular}{l|c|c|c|c}
        \toprule
        \textbf{Cancer} & \textbf{Total} & \textbf{SL} & \textbf{non-SL} & \textbf{Unlabeled} \\
        \midrule
        BRCA    & 2453          & 1443  & 1010          & 151888 \\
        OV      & 805           & 253   & 552           & 151972 \\
        CESC    & 4900          & 144   & 4756          & 150964 \\
        SKCM    & 18407         & 107   & 18300         & 151545 \\
        LUAD    & 6103          & 594   & 5509          & 150944 \\
        \bottomrule
    \end{tabular}
\end{table}

\begin{table}[!htbp]
    \centering
    \caption{Final distribution of classes in ST in the Randomized split experiments. The percentage of final train sets that are reported are averaged over 10 runs.}
    \label{tab:classimbalance}
    \begin{tabular}{l|c|c|c|c}
        \toprule
         \textbf{Cancer} & \textbf{Share majority class (\%)} & \textbf{Share minority class (\%)} \\
         \midrule
         BRCA & 55 $\pm$ 2 & 45 $\pm$ 2 \\
         OV & 69 $\pm$ 3 & 31 $\pm$ 3 \\
         SKCM & 70 $\pm$ 5 & 30 $\pm$ 5 \\
         CESC & 68 $\pm$ 6 & 32 $\pm$ 6 \\
         LUAD & 56 $\pm$ 2 & 44 $\pm$ 2 \\
         \bottomrule
    \end{tabular}
\end{table}

\begin{table}[!htbp]
    \centering
    \caption{Sizes of individual runs for the randomized split experiment after splitting the data and balancing for each set.\label{tab:randomized_split}}
    \begin{tabular}{l|l|c|c|c|c|c|c|c|c|c}
        \toprule%
        \textbf{Run} & \textbf{Cancer} & \textbf{Train} & \textbf{SL} & 
        \textbf{\begin{tabular}[c]{@{}c@{}}non\\SL\end{tabular}} & \textbf{Validation} & \textbf{SL} & \textbf{\begin{tabular}[c]{@{}c@{}}non\\SL\end{tabular}} & \textbf{Test} & \textbf{SL} & \textbf{\begin{tabular}[c]{@{}c@{}}non\\SL\end{tabular}} \\ 
        \midrule
        1       & BRCA  & 1294  & 647   & 647   & 324   & 162   & 162   & 402   & 201   & 201   \\
        2       & BRCA  & 1294  & 647   & 647   & 324   & 162   & 162   & 402   & 201   & 201   \\
        3       & BRCA  & 1294  & 647   & 647   & 324   & 162   & 162   & 402   & 201   & 201   \\
        4       & BRCA  & 1294  & 647   & 647   & 324   & 162   & 162   & 402   & 201   & 201   \\
        5       & BRCA  & 1294  & 647   & 647   & 324   & 162   & 162   & 402   & 201   & 201   \\
        6       & BRCA  & 1294  & 647   & 647   & 324   & 162   & 162   & 402   & 201   & 201   \\
        7       & BRCA  & 1294  & 647   & 647   & 324   & 162   & 162   & 402   & 201   & 201   \\
        8       & BRCA  & 1294  & 647   & 647   & 324   & 162   & 162   & 402   & 201   & 201   \\
        9       & BRCA  & 1294  & 647   & 647   & 324   & 162   & 162   & 402   & 201   & 201   \\
        10      & BRCA  & 1294  & 647   & 647   & 324   & 162   & 162   & 402   & 201   & 201   \\
        1       & OV    & 324   & 162   & 162   & 80    & 40    & 40    & 102   & 51    & 51    \\
        2       & OV    & 324   & 162   & 162   & 80    & 40    & 40    & 102   & 51    & 51    \\
        3       & OV    & 324   & 162   & 162   & 80    & 40    & 40    & 102   & 51    & 51    \\
        4       & OV    & 324   & 162   & 162   & 80    & 40    & 40    & 102   & 51    & 51    \\
        5       & OV    & 324   & 162   & 162   & 80    & 40    & 40    & 102   & 51    & 51    \\
        6       & OV    & 324   & 162   & 162   & 80    & 40    & 40    & 102   & 51    & 51    \\
        7       & OV    & 324   & 162   & 162   & 80    & 40    & 40    & 102   & 51    & 51    \\
        8       & OV    & 324   & 162   & 162   & 80    & 40    & 40    & 102   & 51    & 51    \\
        9       & OV    & 324   & 162   & 162   & 80    & 40    & 40    & 102   & 51    & 51    \\
        10      & OV    & 324   & 162   & 162   & 80    & 40    & 40    & 102   & 51    & 51    \\
        1       & CESC  & 184   & 92    & 92    & 46    & 23    & 23    & 58    & 29    & 29    \\
        2       & CESC  & 184   & 92    & 92    & 46    & 23    & 23    & 58    & 29    & 29    \\
        3       & CESC  & 184   & 92    & 92    & 46    & 23    & 23    & 58    & 29    & 29    \\
        4       & CESC  & 184   & 92    & 92    & 46    & 23    & 23    & 58    & 29    & 29    \\
        5       & CESC  & 184   & 92    & 92    & 46    & 23    & 23    & 58    & 29    & 29    \\
        6       & CESC  & 184   & 92    & 92    & 46    & 23    & 23    & 58    & 29    & 29    \\
        7       & CESC  & 184   & 92    & 92    & 46    & 23    & 23    & 58    & 29    & 29    \\
        8       & CESC  & 184   & 92    & 92    & 46    & 23    & 23    & 58    & 29    & 29    \\
        9       & CESC  & 184   & 92    & 92    & 46    & 23    & 23    & 58    & 29    & 29    \\
        10      & CESC  & 184   & 92    & 92    & 46    & 23    & 23    & 58    & 29    & 29    \\
        1       & SKCM  & 138   & 69    & 69    & 34    & 17    & 17    & 42    & 21    & 21    \\
        2       & SKCM  & 138   & 69    & 69    & 34    & 17    & 17    & 42    & 21    & 21    \\
        3       & SKCM  & 138   & 69    & 69    & 34    & 17    & 17    & 42    & 21    & 21    \\
        4       & SKCM  & 138   & 69    & 69    & 34    & 17    & 17    & 42    & 21    & 21    \\
        5       & SKCM  & 138   & 69    & 69    & 34    & 17    & 17    & 42    & 21    & 21    \\
        6       & SKCM  & 138   & 69    & 69    & 34    & 17    & 17    & 42    & 21    & 21    \\
        7       & SKCM  & 138   & 69    & 69    & 34    & 17    & 17    & 42    & 21    & 21    \\
        8       & SKCM  & 138   & 69    & 69    & 34    & 17    & 17    & 42    & 21    & 21    \\
        9       & SKCM  & 138   & 69    & 69    & 34    & 17    & 17    & 42    & 21    & 21    \\
        10      & SKCM  & 138   & 69    & 69    & 34    & 17    & 17    & 42    & 21    & 21    \\
        1       & LUAD  & 760   & 380   & 380   & 190   & 95    & 95    & 238   & 119   & 119   \\
        2       & LUAD  & 760   & 380   & 380   & 190   & 95    & 95    & 238   & 119   & 119   \\
        3       & LUAD  & 760   & 380   & 380   & 190   & 95    & 95    & 238   & 119   & 119   \\
        4       & LUAD  & 760   & 380   & 380   & 190   & 95    & 95    & 238   & 119   & 119   \\
        5       & LUAD  & 760   & 380   & 380   & 190   & 95    & 95    & 238   & 119   & 119   \\
        6       & LUAD  & 760   & 380   & 380   & 190   & 95    & 95    & 238   & 119   & 119   \\
        7       & LUAD  & 760   & 380   & 380   & 190   & 95    & 95    & 238   & 119   & 119   \\
        8       & LUAD  & 760   & 380   & 380   & 190   & 95    & 95    & 238   & 119   & 119   \\
        9       & LUAD  & 760   & 380   & 380   & 190   & 95    & 95    & 238   & 119   & 119   \\
        10      & LUAD  & 760   & 380   & 380   & 190   & 95    & 95    & 238   & 119   & 119   \\
        \bottomrule
    \end{tabular}
\end{table}

\begin{table}[!htbp]
    \centering
    \caption{Sizes of individual runs for the double holdout experiments\label{tab:double_holdout}}
    \begin{tabular}{l|l|c|c|c|c|c|c|c|c|c}
        \toprule
        \textbf{Run} & \textbf{Cancer} & \textbf{Train} & \textbf{SL} & 
        \textbf{\begin{tabular}[c]{@{}c@{}}non\\SL\end{tabular}} & \textbf{Validation} & \textbf{SL} & \textbf{\begin{tabular}[c]{@{}c@{}}non\\SL\end{tabular}} & \textbf{Test} & \textbf{SL} & \textbf{\begin{tabular}[c]{@{}c@{}}non\\SL\end{tabular}} \\
        \midrule
        1       & BRCA  & 520   & 260   & 260   & 130   & 65    & 65    & 214   & 107   & 107 \\
        2       & BRCA  & 532   & 266   & 266   & 132   & 66    & 66    & 210   & 105   & 105 \\
        3       & BRCA  & 460   & 230   & 230   & 114   & 57    & 57    & 216   & 108   & 108 \\
        4       & BRCA  & 524   & 262   & 262   & 130   & 65    & 65    & 208   & 104   & 104 \\
        5       & BRCA  & 558   & 279   & 279   & 140   & 70    & 70    & 202   & 101   & 101 \\
        6       & BRCA  & 494   & 247   & 247   & 124   & 62    & 62    & 202   & 101   & 101 \\
        7       & BRCA  & 460   & 230   & 230   & 116   & 58    & 58    & 256   & 128   & 128 \\
        8       & BRCA  & 520   & 260   & 260   & 130   & 65    & 65    & 224   & 112   & 112 \\
        9       & BRCA  & 534   & 267   & 267   & 134   & 67    & 67    & 202   & 101   & 101 \\
        10      & BRCA  & 538   & 269   & 269   & 134   & 67    & 67    & 206   & 103   & 103 \\
        1       & OV    & 142   & 71    & 71    & 36    & 18    & 18    & 42    & 21    & 21 \\
        2       & OV    & 142   & 71    & 71    & 36    & 18    & 18    & 50    & 25    & 25 \\
        3       & OV    & 136   & 68    & 68    & 34    & 17    & 17    & 40    & 20    & 20 \\
        4       & OV    & 144   & 72    & 72    & 36    & 18    & 18    & 44    & 22    & 22 \\
        5       & OV    & 148   & 74    & 74    & 36    & 18    & 18    & 44    & 22    & 22 \\
        6       & OV    & 144   & 72    & 72    & 36    & 18    & 18    & 50    & 25    & 25 \\
        7       & OV    & 136   & 68    & 68    & 34    & 17    & 17    & 46    & 23    & 23 \\
        8       & OV    & 152   & 76    & 76    & 38    & 19    & 19    & 44    & 22    & 22 \\
        9       & OV    & 140   & 70    & 70    & 36    & 18    & 18    & 46    & 23    & 23 \\
        10      & OV    & 140   & 70    & 70    & 34    & 17    & 17    & 44    & 22    & 22 \\
        1       & CESC  & 90    & 45    & 45    & 22    & 11    & 11    & 28    & 14    & 14 \\
        2       & CESC  & 90    & 45    & 45    & 22    & 11    & 11    & 30    & 15    & 15 \\
        3       & CESC  & 90    & 45    & 45    & 22    & 11    & 11    & 28    & 14    & 14 \\
        4       & CESC  & 92    & 46    & 46    & 22    & 11    & 11    & 30    & 15    & 15 \\
        5       & CESC  & 90    & 45    & 45    & 22    & 11    & 11    & 28    & 14    & 14 \\
        6       & CESC  & 88    & 44    & 44    & 22    & 11    & 11    & 30    & 15    & 15 \\
        7       & CESC  & 92    & 46    & 46    & 22    & 11    & 11    & 30    & 15    & 15 \\
        8       & CESC  & 88    & 44    & 44    & 22    & 11    & 11    & 28    & 14    & 14 \\
        9       & CESC  & 92    & 46    & 46    & 22    & 11    & 11    & 28    & 14    & 14 \\
        10      & CESC  & 90    & 45    & 45    & 22    & 11    & 11    & 24    & 12    & 12 \\
        1       & SKCM  & 120   & 60    & 60    & 30    & 15    & 15    & 22    & 11    & 11 \\
        2       & SKCM  & 120   & 60    & 60    & 30    & 15    & 15    & 22    & 11    & 11 \\
        3       & SKCM  & 120   & 60    & 60    & 30    & 15    & 15    & 22    & 11    & 11 \\
        4       & SKCM  & 120   & 60    & 60    & 30    & 15    & 15    & 22    & 11    & 11 \\
        5       & SKCM  & 120   & 60    & 60    & 30    & 15    & 15    & 22    & 11    & 11 \\
        6       & SKCM  & 120   & 60    & 60    & 30    & 15    & 15    & 22    & 11    & 11 \\
        7       & SKCM  & 120   & 60    & 60    & 30    & 15    & 15    & 22    & 11    & 11 \\
        8       & SKCM  & 120   & 60    & 60    & 30    & 15    & 15    & 22    & 11    & 11 \\
        9       & SKCM  & 120   & 60    & 60    & 30    & 15    & 15    & 22    & 11    & 11 \\
        10      & SKCM  & 120   & 60    & 60    & 30    & 15    & 15    & 22    & 11    & 11 \\
        1       & LUAD  & 322   & 161   & 161   & 80    & 40    & 40    & 106   & 53    & 53 \\
        2       & LUAD  & 334   & 167   & 167   & 84    & 42    & 42    & 104   & 52    & 52 \\
        3       & LUAD  & 322   & 161   & 161   & 80    & 40    & 40    & 100   & 50    & 50 \\
        4       & LUAD  & 334   & 167   & 167   & 84    & 42    & 42    & 106   & 53    & 53 \\
        5       & LUAD  & 334   & 167   & 167   & 84    & 42    & 42    & 104   & 52    & 52 \\
        6       & LUAD  & 380   & 190   & 190   & 94    & 47    & 47    & 120   & 60    & 60 \\
        7       & LUAD  & 340   & 170   & 170   & 84    & 42    & 42    & 106   & 53    & 53 \\
        8       & LUAD  & 348   & 174   & 174   & 86    & 43    & 43    & 108   & 54    & 54 \\
        9       & LUAD  & 322   & 161   & 161   & 80    & 40    & 40    & 100   & 50    & 50 \\
        10      & LUAD  & 330   & 165   & 165   & 82    & 41    & 41    & 106   & 53    & 53 \\
        \bottomrule
    \end{tabular}
\end{table}

\begin{table}[!htbp]
    \centering
    \caption{Sizes of datasets for the Multiple SL label sources experiments.\label{tab:multiple-experiments}}
    \begin{tabular}{l|l|l|c|c|c|c|c|c|c}
        \toprule
        \textbf{\begin{tabular}[l]{@{}l@{}}Training\\ study\end{tabular}} & 
        \textbf{\begin{tabular}[l]{@{}l@{}}Test\\ study\end{tabular}} & 
        \textbf{Cancer} & \textbf{Train} & \textbf{SL} & \textbf{non-SL} & \textbf{Test} & \textbf{SL} & \textbf{non-SL} & \textbf{Unlabeled} \\
        \midrule
        ISLE & dSL & BRCA & 1509 & 573 & 935 & 960 & 885 & 75 & 151882 \\
        dSL & ISLE & BRCA & 893 & 854 & 39 & 1575 & 590 & 985 & 151882 \\
        ISLE & dSL & LUAD & 4897 & 168 & 4729 & 711 & 372 & 339 & 150944 \\
        dSL & ISLE & LUAD & 711 & 372 & 339 & 4897 & 168 & 4729 & 150944 \\
        EXP2SL & dSL & LUAD & 2676 & 307 & 2369 & 711 & 372 & 339 & 150944 \\
        dSL & EXP2SL & LUAD & 711 & 372 & 339 & 2676 & 307 & 2369 & 150944 \\
        \bottomrule
    \end{tabular}
\end{table}

\begin{table}[!htbp]
    \centering
    \caption{Selected parameters for the randomized split experiments\label{tab:parameters-randomized-split}. For $\mu$, the confidence threshold, the values 0.80, 0.85, 0.90 and 0.95 were tested. For $p$, number of pseudo-labeled samples to add in each iteration of self-training, the values 10, 20 and 50 were tested.}
    \begin{tabular}{l|c|c}
        \toprule
        \textbf{\begin{tabular}[l]{@{}l@{}}Cancer\end{tabular}} & 
        \textbf{\begin{tabular}[c]{@{}c@{}}$p$\end{tabular}} & 
        \textbf{\begin{tabular}[c]{@{}c@{}}$\mu$\end{tabular}} \\
        \midrule 
        BRCA & 0.90 & 20 \\
        OV & 0.85 & 20 \\
        CESC & 0.90 & 10 \\
        SKCM & 0.90 & 10 \\
        LUAD & 0.90 & 10 \\
        \bottomrule
    \end{tabular}
\end{table}

\begin{table}[!htbp]
    \centering
    \caption{Selected parameters for the double holdout experiments\label{tab:parameters-double-holdout}. For $\mu$, the confidence threshold, the values 0.70, 0.75, 0.80, 0.85, 0.90 and 0.95 were tested. For $p$, number of pseudo-labeled samples to add in each iteration of self-training, the values 6, 10 and 20 were tested.}
    \begin{tabular}{l|c|c}
        \toprule
        \textbf{\begin{tabular}[l]{@{}l@{}}Cancer\end{tabular}} & 
        \textbf{\begin{tabular}[c]{@{}c@{}}$p$\end{tabular}} & 
        \textbf{\begin{tabular}[c]{@{}c@{}}$\mu$\end{tabular}} \\
        \midrule 
        BRCA & 0.85 & 6 \\
        OV & 0.80 & 6 \\
        CESC & 0.75 & 6 \\
        SKCM & 0.75 & 20 \\
        LUAD & 0.90 & 10 \\
        \bottomrule
    \end{tabular}
\end{table}

\begin{table}[!htbp]
    \centering
    \caption{Selected parameters for the Multiple SL label sources experiments\label{tab:parameters-multiple-experiments}. For $\mu$, the confidence threshold, the values 0.75, 0.80, 0.85, 0.90 and 0.95 were tested. For $p$, number of pseudo-labeled samples to add in each iteration of self-training, the values 4, 6, 10 and 20 were tested.}
    \begin{tabular}{l|l|l|c|c}
        \toprule
        \textbf{\begin{tabular}[l]{@{}l@{}}Training\\ study\end{tabular}} & 
        \textbf{\begin{tabular}[l]{@{}l@{}}Test\\ study\end{tabular}} & 
        \textbf{\begin{tabular}[l]{@{}l@{}}Cancer\end{tabular}} & 
        \textbf{\begin{tabular}[c]{@{}c@{}}$p$\end{tabular}} & 
        \textbf{\begin{tabular}[c]{@{}c@{}}$\mu$\end{tabular}} \\
        \midrule 
        ISLE & dSL & BRCA & 6 & 0.85 \\
        dSL & ISLE & BRCA & 4 & 0.95 \\
        ISLE & dSL & LUAD & 10 & 0.80 \\
        dSL & ISLE & LUAD & 10 & 0.85 \\
        EXP2SL & dSL & LUAD & 6 & 0.85 \\
        dSL & EXP2SL & LUAD & 10 & 0.80 \\
        \bottomrule
    \end{tabular}
\end{table}

\clearpage
\section{Supplementary Methods}
\subsection*{Details of the double holdout experiment}
To assess the performance of the proposed methods when the train and test set follow different biases, we performed an experiment where the gene pairs in the train and test sets did not have any genes in common. By decoupling the genes in the train set from the test set, we constructed an experiment where the two sets do not originate from the same distribution and do not follow the same sample selection bias. In this experiment, we could evaluate the ability of the methods to transfer knowledge learned on one distribution to data with a different bias. For BRCA, CESC, LUAD, and OV, we divided the set of all individual genes instead of pairs into two sets: a training and a test gene set. Then all pairwise combinations of genes with available SL labels were generated within each set while trying to protect the ratio of samples between the training and test set to 4:1. We generated 10 different runs where in each run, the gene sets were selected randomly. This separation ensured that there was no overlap between the two sets of gene pairs. In contrast, for the SKCM dataset, since the gene \textit{MYC} was dominant and only 60 samples did not contain the \textit{MYC} gene, we constructed the test set always from these pairs without \textit{MYC} gene. Then, for the training set, we used all pairs except those 60 and any other pair that had any gene overlap with these 60 samples.

\clearpage
\bibliography{sample}